\documentclass[
]{ceurart}
\usepackage{float}

\usepackage{listings}
\usepackage{xcolor}
\usepackage{bbm}

\colorlet{punct}{red!60!black}
\definecolor{background}{HTML}{EEEEEE}
\definecolor{delim}{RGB}{20,105,176}
\colorlet{numb}{magenta!60!black}
\lstdefinelanguage{json}{
    basicstyle=\normalfont\ttfamily,
    showstringspaces=false,
    breaklines=true,
    frame=lines,
    backgroundcolor=\color{background},
    literate=
     *{0}{{{\color{numb}0}}}{1}
      {1}{{{\color{numb}1}}}{1}
      {2}{{{\color{numb}2}}}{1}
      {3}{{{\color{numb}3}}}{1}
      {4}{{{\color{numb}4}}}{1}
      {5}{{{\color{numb}5}}}{1}
      {6}{{{\color{numb}6}}}{1}
      {7}{{{\color{numb}7}}}{1}
      {8}{{{\color{numb}8}}}{1}
      {9}{{{\color{numb}9}}}{1}
      {:}{{{\color{punct}{:}}}}{1}
      {,}{{{\color{punct}{,}}}}{1}
      {\{}{{{\color{delim}{\{}}}}{1}
      {\}}{{{\color{delim}{\}}}}}{1}
      {[}{{{\color{delim}{[}}}}{1}
      {]}{{{\color{delim}{]}}}}{1},
}

\usepackage[labelfont=bf, labelsep=period, hypcap=false]{caption}

\begin{document}

\copyrightyear{2023}
\copyrightclause{Copyright for this paper by its authors.
  Use permitted under Creative Commons License Attribution 4.0
  International (CC BY 4.0).}

\conference{IberLEF 2023, September 2023, Jaén, Spain}

\title{A Framework for Identifying Depression on Social Media: MentalRiskES@IberLEF 2023}


\author[1]{Simón Sánchez Viloria}[%
email=simsanch@inf.uc3m.es,
]
\cormark[1]
\address[1]{Universidad Carlos III de Madrid (UC3M),
  Av. Universidad, 30 (edificio Sabatini), 28911 Leganés (Madrid), Spain}

\author[1]{Daniel Peix del Río}[%
email=dpeix@pa.uc3m.es,
]
\fnmark[1]

\author[1]{Rubén Bermúdez Cabo}[%
email=100384003@alumnos.uc3m.es,
]
\fnmark[1]

\author[1]{Guillermo Arturo Arrojo Fuentes}[%
email=100501115@alumnos.uc3m.es,
]
\fnmark[1]

\author[1]{Isabel Segura-Bedmar}[%
orcid=0000-0002-7810-2360,
email=isegura@inf.uc3m.es,
url=https://researchportal.uc3m.es/display/inv25506,
]

\cortext[1]{Corresponding author.}
\fntext[1]{These authors contributed equally.}

\begin{abstract}
  This paper describes our participation in the MentalRiskES task at IberLEF 2023. The task involved predicting the likelihood of an individual experiencing depression based on their social media activity. The dataset consisted of conversations from 175 Telegram users, each labeled according to their evidence of suffering from the disorder. We used a combination of traditional machine learning and deep learning techniques to solve four predictive subtasks: binary classification, simple regression, multiclass classification, and multi-output regression. 
  We approached this by training a model to solve the multi-output regression case and then transforming the predictions to work for the other three subtasks. 
  We compare the performance of two modeling approaches: fine-tuning a BERT-based model directly for the task or using its embeddings as inputs to a linear regressor, with the latter yielding better results. The code to reproduce our results can be found at: \href{https://github.com/simonsanvil/EarlyDepression-MentalRiskEs/tree/main}{\textcolor{blue}{https://github.com/simonsanvil/EarlyDepression-MentalRiskES}}
\end{abstract}


\maketitle

\vspace*{-0.2cm}
\section{Introduction}
\label{sec:intro}

Mental health is a growing concern in our society. According to the World Health Organization (WHO), 1 in 4 people will be affected by mental disorders at some point in their lives \cite{who_mental_2001}. In addition, the COVID-19 pandemic has had a negative impact on the mental health of the general population, with an increase in the number of people suffering from mental disorders \cite{xiong_impact_2020}. 
Thus, it is becoming increasingly important to evaluate the use of new technologies to assess the risk of mental illness and the healthcare needs of the population \cite{losada2017erisk}.



At the same time, social media platforms such as \textit{Telegram} have become a popular way for people to express their feeling and emotions. Telegram is a free, end-to-end encrypted messaging service that allows users to send and receive messages and media files in private chats or groups that can be focused on particular topics and allow any user to observe or actively participate.  These characteristics make Telegram a suitable source for text-mining \cite{dargahi_nobari_analysis_2017}.

With this context, an interesting approach is to use Natural Language Processing (NLP) techniques to analyze the language used by people who suffer from mental illness and discover patterns that can be used to identify them and provide the necessary support. The MentalRiskES task at IberLEF 2023 \cite{MentalRiskES2023} aims to promote the development of NLP solutions specifically for Spanish-speaking social media. They propose three main areas of focus for early-risk detection: eating disorders (Task 1), depression (Task 2), and non-defined disorders (Task 3).

In this work, we present our proposed solution to Task 2 of the 2023 edition of MentalRiskES. This task involves evaluating the likelihood of a Telegram user experiencing depression based on their comments within mental-health-focused groups. The task is split into four predictive subtasks (2a, 2b, 2c, 2d) according to the type of output required. Our main contributions and findings can be then summarized threefold:

\begin{enumerate}
  \item  We conducted experiments using various language models based on BERT \cite{devlin_bert_2019} to solve the task. We found that a RoBERTa model \cite{gutierrez_roberta_2022} that had been previously fine-tuned on a Spanish corpus to identify suicide behavior \cite{roberta-base-bne-finetuned-suicide-es_2023} tended to yield the most accurate results. This suggests that fine-tuning for an intermediate task can improve results for related tasks, which is supported by existing literature \cite{phang_sentence_2019, chang_rethinking_2021}.
  \item Our approach to solving the task consisted of training only with the labels of the regression subtasks (2b, 2d), as we deemed them the most informative. Additionally, we show that you can use the labels of 2d to recover the labels of the other three subtasks. The models trained to target task 2d achieved the best results across all subtasks, even outperforming those that targeted 2b in the simple regression metrics. 
  \item We attempted two different predictive modeling approaches to solve the task using the language model (LM) mentioned above. The first one involved extracting the \textit{sentence embeddings} of the messages of each user and using them as features to train and evaluate classic linear and non-linear machine-learning regressors.
  In the second one, we fine-tuned the LM directly for the subtask.
  The first approach proved advantageous in terms of allowing for quicker, more comprehensive experimentation and resulted in models that achieved the best overall performance when evaluated on the test set.
\end{enumerate}

The rest of the paper is organized as follows: In the next section, we analyze the dataset used for the task (Section \ref{sec:dataset-analysis}). Then, we describe in detail our methodology for training and evaluating the models (Section \ref{sec:methodology}). Finally, we discuss the results obtained (Section \ref{sec:results}) and present our conclusions and future lines of work (Section \ref{sec:conclusions}).

\section{Dataset Analysis} \label{sec:dataset-analysis}

\begin{figure}[!ht]
  \centering
  \includegraphics[width=0.9
  \linewidth]{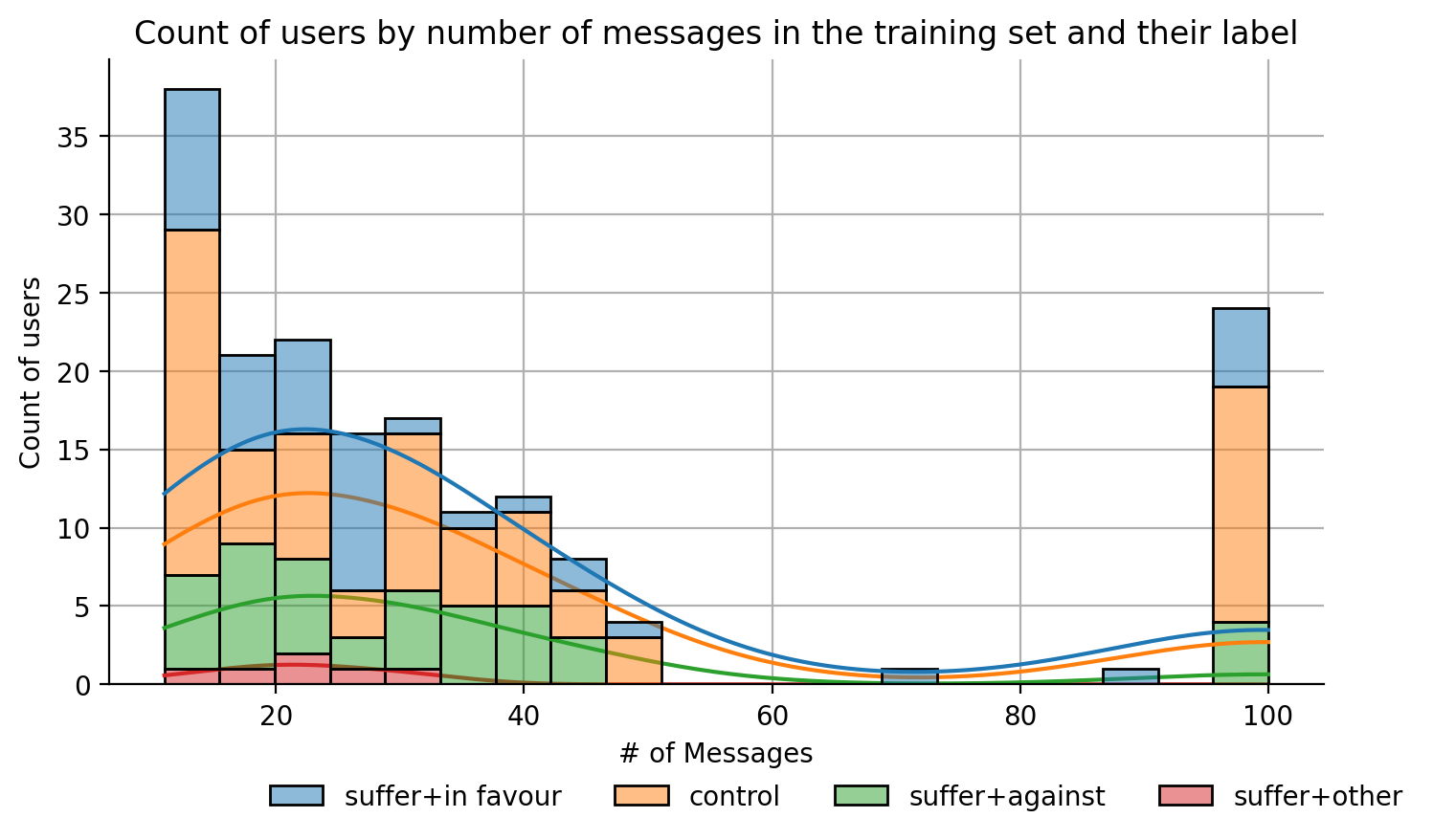}
  \caption{Plot of the count of users by the number of messages in the training set and their assigned label. x-axis: number of messages per user. y-axis: count of users with that amount of messages.}
  \label{fig:fig1}
\end{figure}

The dataset given for the task consisted of a total of \textit{6,248} individual messages from 175 Telegram users, each with a variable number of messages (see figure \ref{fig:fig1}). The annotation process consisted of labeling each user based on the evidence from their conversation history of suffering from depression. Thus, a total of 10 annotators were used for the tasks. Each was asked to assign one of the following four labels to each user:

\clearpage

\begin{itemize}
  \item \textbf{suffer+in favour}: Indicates evidence (from text messages) of the user suffering from depression but is also receptive/willing to help and overcome it.
  \item \textbf{suffer+against}: Indicates evidence of the user suffering from depression but is against receiving or providing help to overcome it.
  \item \textbf{suffer+other}:  Indicates evidence of the user suffering from depression, but there's not enough information to assign them to any further category (against or in favour)
  \item \textbf{control}: Indicates no evidence of the user suffering from depression.
\end{itemize}

\vspace*{-0.5cm}

Furthermore, these labels were represented differently to support each of the four subtasks of MentalRiskES: simple classification (\textit{task 2a}), binary regression (\textit{task 2b}), multiclass classification (\textit{task 2c}), and multi-output regression (\textit{task 2d}). 

In the classification tasks (2a, 2c), the label assigned to each user was the class that obtained the majority vote from the annotators, with the labels being "1" for the "suffer" classes and "0" for the control in the case of task 2a. For the regression tasks (2b, 2d), the values of the labels were presented as numeric probabilities in $[0, 1]$ representing the confidence of the respective class. They were calculated by adding the number of annotators who gave the classification and dividing by 10 (the total number of annotators). For task 2b, this was presented as one number representing the probability of suffering from depression, while for task 2d, each subject label was presented with four numbers representing the probability of each class. Appendix \ref{appendix:dataset-examples} shows examples of how this data was given.


The following figure displays the label distribution for each task in the training set. We can see how over 94 ($\sim$54\%) of users were classified as having depression. Furthermore, there is an imbalance in the labels for the classification tasks due to the "suffer" label being divided into different categories (leading to an over-representation of the "control" label). Additionally, the "suffer+other" category is underrepresented when compared to the other three.

\begin{center}
  \begin{minipage}{\linewidth}
    \centering
    \includegraphics[width=0.74\linewidth]{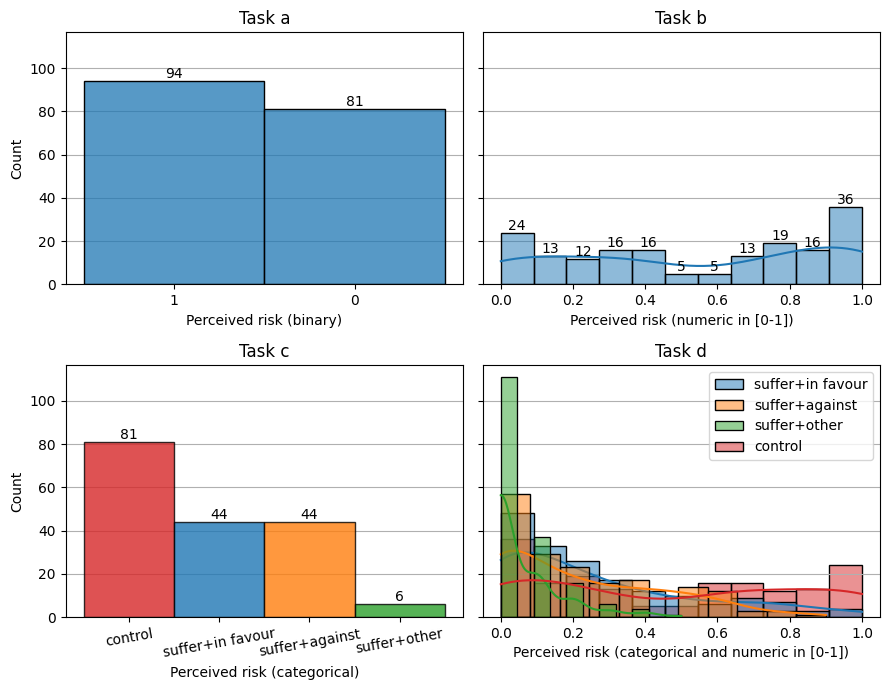}
    \captionof{figure}{Distribution of labels assigned to the users for each task. In tasks a and d: 1 = evidence of suffering from depression. Task d: The sum of the four labels adds up to 1.}
    \label{fig:fig2}
  \end{minipage}
\end{center}


\section{Methodology}
\label{sec:methodology}

We proceeded to evaluate different techniques to solve each of the four subtasks. Two main predictive-modeling approaches were explored: The first one involved fine-tuning a pre-trained language model on each subtask and the second was about training a standard ML regressor using sentence embeddings encoded from the user's messages as features. The following section describes the steps taken for each approach, first describing how the data was pre-processed and later explaining the training and evaluation process done for each subtask.

\subsection{Data Processing and Augmentation}

Independent of the approach taken to train the models, the data was pre-processed and augmented in the same way. The first thing we did was group all the messages by the user they belonged to and concatenated them into a single string, obtaining a total of 175 messages (one per user). This was done to obtain a single representation of each user's conversation history (from which the labels were assigned) to be able to use it as input for the models. 

To prepare for training, the data was split into training and validation sets, leaving a random 26 (15\%) users in the latter for stratified cross-validation, where each set receives the same proportion of samples of each class \cite{kohavi_study_1995}. The stratification was done using the labels of task c to ensure equal representation of the classes in both sets. 

To increase the amount of data available for training and, at the same time, attempt to model early detection (obtaining predictions early on in the lifetime of the message history), we augmented the training set by adding observations that only contained \textit{half} of their messages. This was done by first sorting the messages of each user in the training set by its date and then only taking the first half, the resulting dataset was then appended to the original training set to obtain a new one with twice the number of observations to be used for training.

\subsection{Solving all substaks by solving for regression} \label{sec:solving-for}

By the discussion in section \ref{sec:dataset-analysis}, it should be clear to see that not all labels of the subtasks give the same amount of information about the condition of the subject and the likelihood of predicting it based on the available data. Indeed, it's clear that the probability values of task 2b give more information about confidence in predicting depression than the simple binary labels of task 2a. For the same reasons, the labels of task 2d are more informative than those of task 2c as they give the full probability distribution across the four classes.

Furthermore, we can show that it's possible to use the multi-output regression labels (2d) to recover the labels of the other three subtasks. To illustrate, the multiclass classification labels of task 2c can be recovered by selecting the class in the distribution that has the highest probability. Moreover, we can obtain the labels of task 2a by simply converting these classes into binary (1 for the "suffer" classes and 0 for all others). Lastly, the labels of task 2b can be obtained by summing the probabilities of the "suffer" classes in the distribution. We have confirmed this by applying these modifications to the labels of the training set for task 2d and comparing them to the original labels of the other three tasks.

This observation led us to consider using models that solve for more than one subtask by only training it with the labels of task 2b or 2d. This allowed us to reduce the number of models that had to be trained and focus on solving for a single data modality (regression on $[0,1]$). 

We approached simple regression in a standard way training models, training models to minimize the Mean Squared Error between the output values and the real ones. Additionally, we included the post-processing step of clipping the output predictions of models of this type to the [0,1] range to ensure that they were valid probabilities. 

Multi-output regression using standard machine learning regression, on the other hand, wasn't as trivial as in the simple regression case. The models we worked on didn't support multi-output regression out of the box. The approach we did involve training four regressors for each model, one for each class, and then combining the predictions. We explored two methods for this: training independent regressors or training them in a chain as explained by figure \ref{fig:multi-regression-methods}. The full details of the process are described in appendix \ref{appendix:multi-output-regression}.

Finally, similar to the simple regression case, the predictions of the multi-output models were post-processed by dividing each of the four values by their sum to obtain a vector whose values add up to one. That is, $\hat{y}_i = \frac{\hat{y}_i}{\sum_{i} \hat{y}_i}$ for each $i$-th class. This was done to ensure that the predictions were valid probability distributions over the classes. 
\clearpage

\begin{figure}[h]
  \centering
  \includegraphics[width=0.7\textwidth]{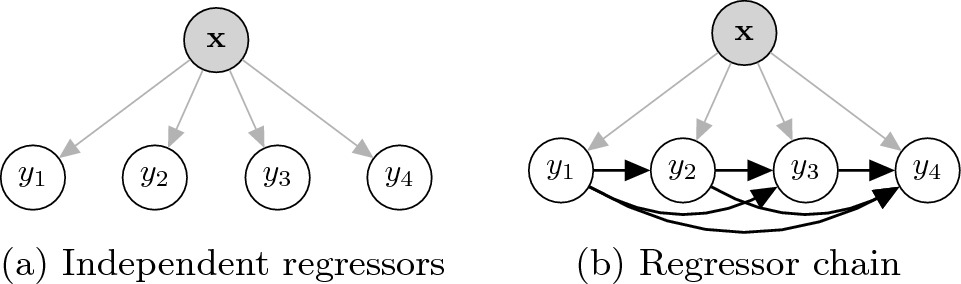}
  \caption{Graphical representation of the two methods used to implement multi-output regression taken from \cite{antonenko_multi-modal_2022}. In (a), the regressors are trained independently with the same input, while in (b) they are trained in a chain with the predictions of the previous ones being passed as features to the next.}
  \label{fig:multi-regression-methods}
\end{figure}
\vspace*{-0.6cm}
\subsection{Modeling Approaches} \label{modeling-approaches}

\subsubsection{Training a regressor with sentence embeddings}

A sentence embedding is a semantically meaningful real-valued vector representation of a sentence, obtained from the outputs of the hidden layers of a language model. The properties of this representation are so that sentences that express similar meanings are mapped (encoded) closer to each other in the vector space \cite{perone_evaluation_2018}. 

In this way, the process of encoding text as numeric vectors can be used directly to extract features for a classifier or regressor, which will try to learn from the semantic information of these encodings to predict the label of their corresponding messages. Note, however, that this approach requires the need to have a pre-trained model to perform this encoding. Furthermore, it assumes that the model will be good enough at capturing the semantic information of the texts given as input, enough for the classifier/regressor to learn from it.

Assuming that this is the case, this approach has the advantage that it is much faster to train these kinds of regressors with regular CPUs, with the most time-consuming part being obtaining the embeddings of the training/evaluation messages, which only has to be done once. However, it is necessary to evaluate different encoding models and different classifiers/regressors (prediction models) to find the best combination for the task at hand.

As such, we conducted experiments using different language models to find the best encoding model. Particularly, we tested three different versions of BERT \cite{devlin_bert_2019} trained with different corpora in Spanish. These versions are described in table \ref{tab:bert_models}. Additionally, we experimented with over 10 different regressors, including Least Squares Linear regression \cite{hoerl_ridge_1970}, Random Forest \cite{breiman_random_2001}, and Gradient Boosting \cite{friedman_greedy_2000}, among others. These models were chosen due to their ease of implementation and the fact that they are commonly used in the literature \cite{scikit-learn}.

The process of training and evaluating these models proceeded then as follows: First, the training set was encoded using the language model and the resulting embeddings were used as features for a regressor. The regressor was then trained using the labels of task 2d (the most informative ones) and the resulting model was used to predict the labels of the validation set. The predictions were then evaluated with the root mean squared error (RMSE). This process was repeated for each combination of language model and regressor. 

 Appendix \ref{appendix:embeddings} contains the results of this experiment. Based on that, \texttt{roberta-suicide-es} was deemed to be the best model for encoding the texts. Additionally, appendix \ref{appendix:embedding-regressor-eval} shows a detailed report of the evaluation of the best regression model with these embeddings.


\begin{table}
  \begin{center}
  \begin{tabular}{ccc}
      \toprule
      Model&Description\\
      \midrule
      \href{https://huggingface.co/PlanTL-GOB-ES/roberta-base-bne}{RoBERTa-base-bne} \cite{gutierrez_roberta_2022} & RoBERTa model \cite{liu_roberta_2019} trained with data from Spain's National Library. \\
      \href{https://huggingface.co/hackathon-somos-nlp-2023/roberta-base-bne-finetuned-suicide-es}{RoBERTa-suicide-es} \cite{roberta-base-bne-finetuned-suicide-es_2023} & RoBERTa-base-bne fine-tuned for suicide detection.\\
      \href{https://huggingface.co/dccuchile/bert-base-spanish-wwm-cased}{BETO} \cite{CaneteCFP2020} & Variant of BERT \cite{devlin_bert_2019} trained with Spanish corpora.\\
      \bottomrule
 \end{tabular}
 \caption{Pre-trained BERT-based models used in our experiments.}
 \label{tab:bert_models}
 \vspace*{-0.85cm}
\end{center}
\end{table}

\vspace*{-1cm}
\subsubsection{Fine-tuning a Language Model for Regression} 

Apart from the approach mentioned above, we also experimented with the pure Deep Learning (DL) approach of taking a language model and fine-tuning it with the labels of the corresponding subtask. The model we fine-tuned was a version of RoBERTa pre-trained for detecting suicidal behavior from texts in Spanish \cite{roberta-base-bne-finetuned-suicide-es_2023}. We chose this model due to the fact of having been trained previously for a task that shares similar characteristics to ours. Intermediate fine-tuning has been proven to improve the results of downstream tasks by prior literature \cite{phang_sentence_2019,chang_rethinking_2021}.

The HuggingFace Transformers \cite{wolf_transformers_2020} and Pytorch \cite{NEURIPS2019_9015} libraries in Python were utilized for loading the model weights and implementing the training loop. We changed the head of the pre-trained model to a linear layer consisting of output dimension 1 for simple regression or dimension 4 for multi-output regression. The models were trained using an NVIDIA T4 GPU for a total of 30 epochs, where the weights of the pre-trained model remained fully frozen for the first half and then were progressively unfrozen each epoch after that as in \cite{liu_improving_2023}.

\begin{table*}[h]
  \begin{tabular}{ccl}
    \toprule
    Hyperparameters&Value\\
    \midrule
    Optimizer&AdamW\\
    Learning rate&$1e^{-5}$\\
    Max Tokens&1024\\
    Num Epochs&30\\
    Batch Size&1\\
    \bottomrule
  \end{tabular}
  \caption{Hyperparameters for fine-tuning a RoBERTa model for regression tasks.}
  \vspace*{-0.3cm}
  \label{tab:hyper_fine-tuning}
\end{table*}

We used an Adam Optimizer with Mean-Squared Error (MSE) for the simple regression models and a Cross-Entropy loss function for multi-regression (since the labels consisted of numeric probabilities). Furthermore, since the output for task 2d consisted of a probability distribution over the four classes, we experimented with a custom loss function that adds a term to the standard cross-entropy loss to penalize outputs whose sum is different from one. However, this did not improve the results empirically as compared with simply normalizing the outputs of the predictions after inference. The formula of this loss is shown in equation \ref{eq:custom-loss}. Other hyperparameters are shown in table \ref{tab:hyper_fine-tuning}.

\begin{equation}
  \mathcal{L}_{\text{custom}} 
  = \mathcal{L}_{cross-entropy} + \epsilon(1 - \sum_{i \in [1,4]} \hat{y}_i)^2
  \label{eq:custom-loss}
\end{equation}

In the equation above, $\hat{y}$ is the output of the model, $y$ is the target label, $\epsilon$ is a hyperparameter that controls the weight of the penalty term, and $y_i$ is the $i$-th element of the target label. 

\section{Results} \label{sec:results}

Using the approaches mentioned in the prior section, we came up with different models to solve the four subtasks of Task 2 of MentalRiskES. The results in this section are obtained from selecting the best-performing models after evaluating the different approaches and hyperparameters on the validation set. The final predictions were obtained from a test set of messages from 149 subjects never observed during the training process and evaluated against the task's true labels. 

In the tables below, we report the relevant metrics obtained for each subtask and compare them against the ones obtained from baseline models provided by the organizers of the competition. In particular, we report both \textit{absolute} metrics, obtained after observing all the messages of each subject, and \textit{early detection} metrics, obtained after incrementally observing the messages across several rounds. Additionally, table \ref{tab:carbon_emissions} displays the inference-time $\text{CO}_2$ emissions and energy consumption of each model, based on computing their \textit{absolute} predictions on the test set. These values were estimated using the \textit{codecarbon} python library \cite{codecarbon}.

For the absolute metrics, we show the accuracy, precision, recall, and F1 scores for the classification tasks (2a and 2c) and the root mean squared error (RMSE) and coefficient of determination ($R^2$) for the regression tasks (2b and 2d). The early detection metrics include the \textit{early-risk detection} metric (erde) computed after observing different rounds of messages as well as other metrics (more details are provided in the competition guidelines \cite{MentalRiskES2023}). 

The metrics are shown along with the name of the model used to obtain them. The models are named as follows: \textit{[task name]\_[model name]\_[approach]}. For example, \textit{task2b\_roberta-suicide-es\_fine-tuning} refers to the model trained with the task 2b (binary classification) labels by fine-tuning the Roberta model pre-trained for suicide detection. The "\textit{approach}" can be either \textit{embeddings} or \textit{fine-tuning} for the two approaches described in section \ref{sec:methodology}. 

Furthermore, all ML regressors trained with embeddings as features were Ridge regressors, and all embeddings were obtained using \texttt{roberta-suicide-es} encodings as this combination yielded the best results in the evaluation set. The \textit{embeddings} approaches for task 2d also include the multi-regression method used (\textit{ind} indicating that independent regressors were used and \textit{chain} for chained regressors).

\subsection{Results for task 2a: binary classification}

\begin{table}[h]
  \vspace*{-0.3cm}
  \centering
  \caption{Task A absolute Metric Results\\\footnotesize Ranked by Macro F1.}
  \label{tab:task_a_absolute_metrics}
  \begin{tabular}{>{\bfseries}lrrrr}
  \toprule
  {} &  accuracy &  macro\_precision &  macro\_recall &  macro\_f1 \\
  \midrule
  2d\_roberta\_embeddings\_ind    &     \textbf{0.705} &            \textbf{0.717} &         \textbf{0.727} &     \textbf{0.703} \\
   \normalfont BaseLine - Roberta Large &     0.698 &            0.759 &         0.718 &     0.690 \\
   2d\_roberta\_embeddings\_chain   &     \textbf{0.691} &            \textbf{0.711} &         \textbf{0.755} &     \textbf{0.682} \\
  2b\_roberta\_embeddings             &     \textbf{0.691} &            \textbf{0.713} &         \textbf{0.764} &     \textbf{0.681} \\
   2d\_roberta-suicide-es\_fine-tuning            &     \textbf{0.671} &            \textbf{0.695} &         \textbf{0.764} &     \textbf{0.655} \\
   \normalfont BaseLine - Deberta       &     0.664 &            0.788 &         0.691 &     0.642 \\
   2b\_roberta-suicide-es\_fine-tuning            &     \textbf{0.638} &            \textbf{0.663} &         \textbf{0.735} &     \textbf{0.616} \\
   \normalfont BaseLine - Roberta Base  &     0.631 &            0.744 &         0.658 &     0.605 \\
  \bottomrule
  \end{tabular}
\end{table}

\begin{table}[h]
  \centering
  \caption{Task A early-detection Metric Results\\\footnotesize Ranked by ERDE30.}
  \label{tab:task_a_early-detection_metrics}
  \begin{tabular}{lrrrrr}
  \toprule
  {} &  erde30 &  erde5 &  latency\_tp &  latency\_weighted\_f1 &  speed \\
  \midrule
  \textbf{2d\_roberta-suicide-es\_fine-tuning}              &   \textbf{0.013} &  \textbf{0.284} &       \textbf{3.000} &                \textbf{0.716} &  \textbf{0.982} \\
  \textbf{2b\_roberta\_embeddings}               &   \textbf{0.020} &  \textbf{0.286} &       \textbf{3.000} &                \textbf{0.725} &  \textbf{0.982} \\
  \textbf{2b\_roberta-suicide-es\_fine-tuning}              &   \textbf{0.020} &  \textbf{0.208} &       \textbf{2.000} &                \textbf{0.700} &  \textbf{0.991} \\
  \textbf{2d\_roberta\_embeddings\_chain}     &   \textbf{0.027} &  \textbf{0.283} &       \textbf{3.000} &                \textbf{0.722} &  \textbf{0.982} \\
  \textbf{2d\_roberta\_embeddings\_ind}      &   \textbf{0.067} &  \textbf{0.296} &       \textbf{3.000} &                \textbf{0.712} &  \textbf{0.982} \\
  BaseLine - Deberta                   &   0.153 &  0.303 &       2.000 &                0.719 &  0.984 \\
  BaseLine - Roberta Large             &   0.159 &  0.290 &       4.000 &                0.704 &  0.951 \\
  BaseLine - Roberta Base              &   0.176 &  0.342 &       4.000 &                0.671 &  0.951 \\
  \bottomrule
  \end{tabular}
\end{table}


\subsection{Results for task 2b: Simple Regression}

\begin{table}[h]
  \vspace*{-0.3cm}
  \centering
  \caption{Task B absolute Metric Results\\\footnotesize Ranked by RMSE.}
  \label{tab:task_b_absolute_metrics}
  \begin{tabular}{lrr}
  \toprule
  {} &  RMSE &  r2 \\
  \midrule
  \textbf{2d\_roberta\_embeddings\_chain}   & \textbf{0.241} &       \textbf{0.591} \\
  \textbf{2b\_roberta\_embeddings}             & \textbf{0.244} &       \textbf{0.581} \\
  \textbf{2d\_roberta\_embeddings\_ind}    & \textbf{0.259} &       \textbf{0.526} \\
  BaseLine - Roberta Base  & 0.277 &       0.770 \\
  \textbf{2d\_roberta-suicide-es\_fine-tuning}            & \textbf{0.304} &       \textbf{0.349} \\
  \textbf{2b\_roberta-suicide-es\_fine-tuning}            & \textbf{0.311} &       \textbf{0.317} \\
  BaseLine - Deberta       & 0.339 &       0.683 \\
  BaseLine - Roberta Large & 0.390 &       0.503 \\
  \bottomrule
  \end{tabular}
\end{table}

\begin{table}[h]
  \centering
  \caption{Task B Ranking-based  Results: \footnotesize Ranked by the p@30 metric}
  \label{tab:task_b_ranking_metrics}
  \begin{tabular}{lrrrr}
  \toprule
  {} &  p@10 &  p@20 &  p@30 &   p@5 \\
  \midrule
  BaseLine - Roberta Base  & 0.800 & 0.700 & 0.567 & 0.600 \\
  BaseLine - Deberta       & 0.600 & 0.550 & 0.567 & 0.800 \\
  BaseLine - Roberta Large & 0.500 & 0.550 & 0.567 & 0.400 \\
  \textbf{2b\_roberta-suicide-es\_fine-tuning}  & \textbf{0.700} & \textbf{0.700} & \textbf{0.533} & \textbf{1.000} \\
  \textbf{2b\_roberta\_embeddings}   & \textbf{0.800} & \textbf{0.450} & \textbf{0.367} & \textbf{0.800} \\
  \textbf{2d\_roberta\_embeddings\_ind}   & \textbf{0.700} & \textbf{0.350} & \textbf{0.233} & \textbf{0.600} \\
  \textbf{2d\_roberta\_embeddings\_chain}   & \textbf{0.200} & \textbf{0.150} & \textbf{0.133} & \textbf{0.000} \\
  \textbf{2d\_roberta-suicide-es\_fine-tuning}   & \textbf{0.200} & \textbf{0.100} & \textbf{0.100} & \textbf{0.200} \\
  \bottomrule
  \end{tabular}
\end{table}

\clearpage
\subsection{Results for task 2c: Multiclass Classification} \label{sec:multi-output-results}

\begin{table}[h]
  \centering
  \caption{Task C absolute Metric Results\\\footnotesize Ranked by Macro F1.}
  \label{tab:task_c_absolute_metrics}
  \begin{tabular}{lrrrr}
  \toprule
  {} &  accuracy &  macro\_precision &  macro\_recall &  macro\_f1 \\
  \midrule
  \textbf{2d\_roberta-suicide-es\_fine-tuning}            &     \textbf{0.517} &            \textbf{0.446} &         \textbf{0.435} &     \textbf{0.395} \\
  \textbf{2d\_roberta\_embeddings\_ind}    &     \textbf{0.557} &            \textbf{0.429} &         \textbf{0.395} &     \textbf{0.394} \\
  \textbf{2d\_roberta\_embeddings\_chain}   &     \textbf{0.530} &            \textbf{0.437} &         \textbf{0.418} &     \textbf{0.392} \\
  BaseLine - Roberta Large &     0.483 &            0.389 &         0.378 &     0.360 \\
  BaseLine - Deberta       &     0.456 &            0.395 &         0.344 &     0.293 \\
  BaseLine - Roberta Base  &     0.356 &            0.380 &         0.335 &     0.274 \\
  \bottomrule
  \end{tabular}
  \end{table}


\begin{table}[h]
  \centering
  \caption{Task C early-detection Metric Results:
      \\\footnotesize Ranked by ERDE30.}
  \label{tab:task_c_early-detection_metrics}
  \begin{tabular}{lrrrrr}
  \toprule
  {} &  erde30* &  erde5 &  latency\_tp &  latency\_weighted\_f1 &  speed \\
  \midrule
  \textbf{2d\_roberta\_embeddings\_chain}   &   \textbf{0.157} &  \textbf{0.284} &       \textbf{3.000} &                \textbf{0.718} &  \textbf{0.982} \\
  \textbf{2d\_roberta-suicide-es\_fine-tuning}            &   \textbf{0.159} &  \textbf{0.285} &       \textbf{3.000} &                \textbf{0.712} &  \textbf{0.982} \\
  \textbf{2d\_roberta\_embeddings\_ind}    &   \textbf{0.172} &  \textbf{0.297} &       \textbf{3.000} &                \textbf{0.708} &  \textbf{0.982} \\
  BaseLine - Deberta       &   0.190 &  0.330 &       2.000 &                0.695 &  0.984 \\
  BaseLine - Roberta Base  &   0.206 &  0.307 &       2.000 &                0.659 &  0.984 \\
  BaseLine - Roberta Large &   0.232 &  0.283 &       2.000 &                0.652 &  0.984 \\
  \bottomrule
  \end{tabular}
\end{table}

\subsection{Results for task 2d: Multi-output Regression.}

\begin{table}[h]
  \centering
  \caption{Task D absolute Metric Results.\\
  \mbox{\footnotesize Ranked by mean RMSE. Labels are shortened as:
  sf = suffer+in favour, sa = suffer+against, so = suffer+other, c =control}}
  \label{tab:task_d_absolute_metrics}
  \scalebox{0.75}{
  \begin{tabular}{lrrrrrrrrrr}
  \toprule
  {} &  rmse mean* &  rmse sf &  rmse sa &  rmse so &  rmse c &  r2 mean &  r2 sf &  r2 sa &  r2 so &  r2 c \\
  \midrule
  \textbf{2d\_roberta\_embeddings\_chain}   &      \textbf{0.180} &                  \textbf{0.179} &                \textbf{0.191} &              \textbf{0.111} &         \textbf{0.241} &    \textbf{0.355} &                \textbf{0.544} &              \textbf{0.217} &            \textbf{0.069} &       \textbf{0.590} \\
  \textbf{2d\_roberta\_embeddings\_ind}    &      \textbf{0.187} &                  \textbf{0.181} &                \textbf{0.192} &              \textbf{0.114} &         \textbf{0.259} &    \textbf{0.320} &                \textbf{0.532} &              \textbf{0.208} &            \textbf{0.012} &       \textbf{0.526} \\
  \textbf{2d\_roberta-suicide-es\_fine-tuning}            &      \textbf{0.222} &                  \textbf{0.212} &                \textbf{0.230} &              \textbf{0.143} &         \textbf{0.304} &    \textbf{0.006} &                \textbf{0.358} &             \textbf{-0.144} &           \textbf{-0.538} &       \textbf{0.349} \\
  BaseLine - Deberta       &      0.232 &                  0.246 &                0.250 &              0.125 &         0.306 &    0.484 &                0.661 &              0.295 &            0.260 &       0.721 \\
  BaseLine - Roberta Base  &      0.410 &                  0.547 &                0.272 &              0.235 &         0.585 &   -0.145 &               -0.496 &              0.355 &            0.185 &      -0.624 \\
  BaseLine - Roberta Large &      0.437 &                  0.682 &                0.312 &              0.158 &         0.598 &   -0.209 &               -0.678 &              0.890 &            0.059 &      -0.306 \\
  \bottomrule
  \end{tabular}
  }
\end{table}

\clearpage

\begin{table}[h]
  \centering
  \caption{Task D ranking Metric Results: \footnotesize Ranked by p@30}
  \label{tab:task_d_ranking_metrics}
  \begin{tabular}{lrrrrr}
  \toprule
  {} &  p@10 &  p@20 &  p@30 &   p@5 &  p@50 \\
  \midrule
  BaseLine - Deberta       & 0.300 & 0.338 & 0.350 & 0.250 & 0.250 \\
  \textbf{2d\_roberta\_embeddings\_ind}    & \textbf{0.300} & \textbf{0.300} & \textbf{0.292} & \textbf{0.600} & \textbf{0.280} \\
  BaseLine - Roberta Large & 0.275 & 0.263 & 0.275 & 0.350 & 0.350 \\
  \textbf{2d\_roberta\_embeddings\_chain}   & \textbf{0.275} & \textbf{0.275} & \textbf{0.250} & \textbf{0.550} & \textbf{0.240} \\
  BaseLine - Roberta Base  & 0.300 & 0.225 & 0.192 & 0.250 & 0.250 \\
  \textbf{2d\_roberta-suicide-es\_fine-tuning}            & \textbf{0.075} & \textbf{0.113} & \textbf{0.167} & \textbf{0.150} & \textbf{0.145} \\
  \bottomrule
  \end{tabular}
\end{table}

\subsection{Carbon Emissions} \label{carbon-emissions}

\begin{table}[h]
\vspace{-0.35cm}
\centering
  \caption{Estimated $\text{CO}_2$ emissions of each model from predicting all messages on the test set (absolute).\footnotesize\\Estimations were obtained with the \texttt{codecarbon} python library \cite{codecarbon} using a Macbook Pro (2021) w/ M1 Pro and 16GB of RAM for inference. The models with the lowest emissions are highlighted in bold.}
\scalebox{0.93}{
\begin{tabular}{lrrrr}
\toprule
{} &  duration (secs) &  emissions (kgCO2eq) &  cpu\_energy &  ram\_energy \\
model                             &           &            &             &             \\
\midrule
\textbf{2b\_roberta-suicide-es\_fine-tuning} &  \textbf{5.74} &   \textbf{1.56e-06} &    7.97e-06 &    2.53e-07 \\
\textbf{2d\_roberta-suicide-es\_fine-tuning} & \textbf{7.393} &   \textbf{2.00e-06} &    1.03e-05 &    2.58e-07 \\
2b\_roberta\_embeddings             &  23.287 &   6.70e-06 &    3.23e-05 &    2.94e-06 \\
2d\_roberta\_embeddings\_ind         &  23.976 &   6.94e-06 &    3.33e-05 &    3.25e-06 \\
2d\_roberta\_embeddings\_chain       &  23.721 &   7.14e-06 &    3.29e-05 &    4.63e-06 \\
\bottomrule
\end{tabular}
}
\label{tab:carbon_emissions}
\end{table}
\vspace{-0.35cm}

\section{Conclusions}
\label{sec:conclusions}

The results show that the approaches considered in this work were successful at modeling each of the predictive subtasks, with at least one of our models outperforming the baselines in most cases. We can make the following observations:

\begin{itemize}
  \item The best-performing approach across all tasks seems to be the one that uses the embeddings of the messages as input to a multi-output regression model (task 2d). At least one model trained with this approach reached the top ranking for tasks 2a, 2b, and 2d absolute ranking metrics and outperformed the baseline absolute metrics across all tasks. \item Most notably, the regression method that uses multi-output chained regressors obtained the best metrics for task 2d across all models, outperforming the fine-tuning approach by over 20\% in the absolute metrics and reaching the second highest spot in the early-risk metrics for this task.
  \item Models trained for multi-output regression perform very well for binary classification and simple regression tasks, even outperforming the models trained for simple regression targets in their own subtask. This suggests that using one model to solve for multiple targets was indeed a good approach to this problem.
  \item The models obtained with a pure DL approach from fine-tuning a RoBERTa model are estimated to produce over 3-4x \textit{less} emissions at inference time than the hybrid approach from training linear regressors on sentence embeddings. This gap is likely because the fine-tuning approach requires less computation at inference time than the hybrid approach, which requires the computation of the sentence embeddings before feeding them to multiple regressors, while the fine-tuning approach is made in one forward pass.
\end{itemize}

Another finding we can conclude from these insights is that while our models achieve great results in the absolute ranking metrics, they do not perform as well for the metrics that assess early-risk performance. In our work, we did not model explicitly for an early detection scenario; we only added information about prior messages through data augmentation. This limitation means our models may not perform as well in real-world situations where we aim to detect signs of depression in a conversation early on.

Thus, it may be important to explore different training approaches to improve the performance of early-risk detection. This might include directly employing online learning to predict and update the model as new messages come in or incorporating an ensemble of models to make independent decisions about a message's risk level and combining them for a final decision (as seen in [23]). Additionally, we may also look into more efficient implementations of the hybrid approach to minimize the disparity in emissions compared to pure DL models. These improvements are crucial when considering the deployment of our models in real-world situations and will be the focus of future work.


  


\bibliography{sample-ceur}

\begin{thebibliography}{23}
\expandafter\ifx\csname natexlab\endcsname\relax\def\natexlab#1{#1}\fi
\providecommand{\url}[1]{\texttt{#1}}
\providecommand{\href}[2]{#2}
\providecommand{\path}[1]{#1}
\providecommand{\DOIprefix}{doi:}
\providecommand{\ArXivprefix}{arXiv:}
\providecommand{\URLprefix}{URL: }
\providecommand{\Pubmedprefix}{pmid:}
\providecommand{\doi}[1]{\href{http://dx.doi.org/#1}{\path{#1}}}
\providecommand{\Pubmed}[1]{\href{pmid:#1}{\path{#1}}}
\providecommand{\bibinfo}[2]{#2}
\ifx\xfnm\relax \def\xfnm[#1]{\unskip,\space#1}\fi
\bibitem[{{World Health Organization}(2001)}]{who_mental_2001}
\bibinfo{author}{{World Health Organization}}, \bibinfo{title}{The {World}
  {Health} {Report} 2001: {Mental} {Disorders} affect one in four people},
  \bibinfo{year}{2001}. \URLprefix
  \url{https://www.who.int/news/item/28-09-2001-the-world-health-report-2001-mental-disorders-affect-one-in-four-people}.
\bibitem[{Xiong et~al.(2020)Xiong, Lipsitz, Nasri, Lui, Gill, Phan, Chen-Li,
  Iacobucci, Ho, Majeed, and McIntyre}]{xiong_impact_2020}
\bibinfo{author}{J.~Xiong}, \bibinfo{author}{O.~Lipsitz},
  \bibinfo{author}{F.~Nasri}, \bibinfo{author}{L.~M.~W. Lui},
  \bibinfo{author}{H.~Gill}, \bibinfo{author}{L.~Phan},
  \bibinfo{author}{D.~Chen-Li}, \bibinfo{author}{M.~Iacobucci},
  \bibinfo{author}{R.~Ho}, \bibinfo{author}{A.~Majeed}, \bibinfo{author}{R.~S.
  McIntyre},
\newblock \bibinfo{title}{Impact of {COVID}-19 pandemic on mental health in the
  general population: {A} systematic review},
\newblock \bibinfo{journal}{Journal of Affective Disorders}
  \bibinfo{volume}{277} (\bibinfo{year}{2020}) \bibinfo{pages}{55--64}.
  \URLprefix
  \url{https://www.sciencedirect.com/science/article/pii/S0165032720325891}.
  \DOIprefix\doi{10.1016/j.jad.2020.08.001}.
\bibitem[{Losada et~al.(2017)Losada, Crestani, and Parapar}]{losada2017erisk}
\bibinfo{author}{D.~E. Losada}, \bibinfo{author}{F.~Crestani},
  \bibinfo{author}{J.~Parapar},
\newblock \bibinfo{title}{erisk 2017: Clef lab on early risk prediction on the
  internet: experimental foundations},
\newblock in: \bibinfo{booktitle}{Experimental IR Meets Multilinguality,
  Multimodality, and Interaction: 8th International Conference of the CLEF
  Association, CLEF 2017, Dublin, Ireland, September 11--14, 2017, Proceedings
  8}, \bibinfo{organization}{Springer}, \bibinfo{year}{2017}, pp.
  \bibinfo{pages}{346--360}.
\bibitem[{Dargahi~Nobari et~al.(2017)Dargahi~Nobari, Reshadatmand, and
  Neshati}]{dargahi_nobari_analysis_2017}
\bibinfo{author}{A.~Dargahi~Nobari}, \bibinfo{author}{N.~Reshadatmand},
  \bibinfo{author}{M.~Neshati},
\newblock \bibinfo{title}{Analysis of {Telegram}, {An} {Instant} {Messaging}
  {Service}},
\newblock in: \bibinfo{booktitle}{Proceedings of the 2017 {ACM} on {Conference}
  on {Information} and {Knowledge} {Management}}, {CIKM} '17,
  \bibinfo{publisher}{Association for Computing Machinery},
  \bibinfo{address}{New York, NY, USA}, \bibinfo{year}{2017}, pp.
  \bibinfo{pages}{2035--2038}. \URLprefix
  \url{https://dl.acm.org/doi/10.1145/3132847.3133132}.
  \DOIprefix\doi{10.1145/3132847.3133132}.
\bibitem[{Mármol-Romero et~al.(2023)Mármol-Romero, Moreno-Muñoz, Plaza-del
  Arco, Martín-Valdivia, Ureña-López, and Montejo-Ráez}]{MentalRiskES2023}
\bibinfo{author}{A.~M. Mármol-Romero}, \bibinfo{author}{A.~Moreno-Muñoz},
  \bibinfo{author}{F.~M. Plaza-del Arco}, \bibinfo{author}{M.~T.
  Martín-Valdivia}, \bibinfo{author}{L.~A. Ureña-López},
  \bibinfo{author}{A.~Montejo-Ráez},
\newblock \bibinfo{title}{Overview of {M}entalrisk{ES} at {I}ber{LEF} 2023:
  {E}arly {D}etection of {M}ental {D}isorders {R}isk in {S}panish},
\newblock \bibinfo{journal}{Procesamiento del Lenguaje Natural}
  \bibinfo{volume}{71} (\bibinfo{year}{2023}).
\bibitem[{Devlin et~al.(2019)Devlin, Chang, Lee, and
  Toutanova}]{devlin_bert_2019}
\bibinfo{author}{J.~Devlin}, \bibinfo{author}{M.-W. Chang},
  \bibinfo{author}{K.~Lee}, \bibinfo{author}{K.~Toutanova},
  \bibinfo{title}{{BERT}: {Pre}-training of {Deep} {Bidirectional}
  {Transformers} for {Language} {Understanding}}, \bibinfo{year}{2019}.
  \URLprefix \url{http://arxiv.org/abs/1810.04805}.
  \DOIprefix\doi{10.48550/arXiv.1810.04805}, \bibinfo{note}{arXiv:1810.04805
  [cs]}.
\bibitem[{Fandiño et~al.(2022)Fandiño, Estapé, Pàmies, Palao, Ocampo,
  Carrino, Oller, Penagos, Agirre, and Villegas}]{gutierrez_roberta_2022}
\bibinfo{author}{A.~G. Fandiño}, \bibinfo{author}{J.~A. Estapé},
  \bibinfo{author}{M.~Pàmies}, \bibinfo{author}{J.~L. Palao},
  \bibinfo{author}{J.~S. Ocampo}, \bibinfo{author}{C.~P. Carrino},
  \bibinfo{author}{C.~A. Oller}, \bibinfo{author}{C.~R. Penagos},
  \bibinfo{author}{A.~G. Agirre}, \bibinfo{author}{M.~Villegas},
\newblock \bibinfo{title}{Maria: Spanish language models},
\newblock \bibinfo{journal}{Procesamiento del Lenguaje Natural}
  \bibinfo{volume}{68} (\bibinfo{year}{2022}). \URLprefix
  \url{https://upcommons.upc.edu/handle/2117/367156#.YyMTB4X9A-0.mendeley}.
  \DOIprefix\doi{10.26342/2022-68-3}.
\bibitem[{Padial and Gómez(2023)}]{roberta-base-bne-finetuned-suicide-es_2023}
\bibinfo{author}{D.~L. Padial}, \bibinfo{author}{D.~Gómez},
  \bibinfo{title}{hackathon-somos-nlp-2023 -
  roberta-base-bne-finetuned-suicide-es· {Hugging} {Face}},
  \bibinfo{year}{2023}. \URLprefix
  \url{https://huggingface.co/hackathon-somos-nlp-2023/roberta-base-bne-finetuned-suicide-es}.
\bibitem[{Phang et~al.(2019)Phang, Févry, and Bowman}]{phang_sentence_2019}
\bibinfo{author}{J.~Phang}, \bibinfo{author}{T.~Févry}, \bibinfo{author}{S.~R.
  Bowman}, \bibinfo{title}{Sentence {Encoders} on {STILTs}: {Supplementary}
  {Training} on {Intermediate} {Labeled}-data {Tasks}}, \bibinfo{year}{2019}.
  \URLprefix \url{http://arxiv.org/abs/1811.01088}.
  \DOIprefix\doi{10.48550/arXiv.1811.01088}, \bibinfo{note}{arXiv:1811.01088
  [cs]}.
\bibitem[{Chang and Lu(2021)}]{chang_rethinking_2021}
\bibinfo{author}{T.-Y. Chang}, \bibinfo{author}{C.-J. Lu},
  \bibinfo{title}{Rethinking {Why} {Intermediate}-{Task} {Fine}-{Tuning}
  {Works}}, \bibinfo{year}{2021}. \URLprefix
  \url{http://arxiv.org/abs/2108.11696}, \bibinfo{note}{arXiv:2108.11696 [cs]}.
\bibitem[{Kohavi(1995)}]{kohavi_study_1995}
\bibinfo{author}{R.~Kohavi},
\newblock \bibinfo{title}{A {Study} of {Cross}-{Validation} and {Bootstrap} for
  {Accuracy} {Estimation} and {Model} {Selection}},
\newblock in: \bibinfo{booktitle}{IJCAI'95: Proceedings of the 14th
  international joint conference on Artificial intelligence},
  \bibinfo{year}{1995}, pp. \bibinfo{pages}{1137--1143}. \URLprefix
  \url{https://www.semanticscholar.org/paper/A-Study-of-Cross-Validation-and-Bootstrap-for-and-Kohavi/8c70a0a39a686bf80b76cb1b77f9eef156f6432d}.
\bibitem[{Antonenko and Read(2022)}]{antonenko_multi-modal_2022}
\bibinfo{author}{E.~Antonenko}, \bibinfo{author}{J.~Read},
\newblock \bibinfo{title}{Multi-modal {Ensembles} of {Regressor} {Chains}
  for {Multi}-output {Prediction}},
\newblock in: \bibinfo{editor}{T.~Bouadi}, \bibinfo{editor}{E.~Fromont},
  \bibinfo{editor}{E.~Hüllermeier} (Eds.), \bibinfo{booktitle}{Advances in
  {Intelligent} {Data} {Analysis} {XX}}, Lecture {Notes} in {Computer}
  {Science}, \bibinfo{publisher}{Springer International Publishing},
  \bibinfo{address}{Cham}, \bibinfo{year}{2022}, pp. \bibinfo{pages}{1--13}.
  \DOIprefix\doi{10.1007/978-3-031-01333-1_1}.
\bibitem[{Perone et~al.(2018)Perone, Silveira, and
  Paula}]{perone_evaluation_2018}
\bibinfo{author}{C.~S. Perone}, \bibinfo{author}{R.~Silveira},
  \bibinfo{author}{T.~S. Paula}, \bibinfo{title}{Evaluation of sentence
  embeddings in downstream and linguistic probing tasks}, \bibinfo{year}{2018}.
  \URLprefix \url{http://arxiv.org/abs/1806.06259},
  \bibinfo{note}{arXiv:1806.06259 [cs] version: 1}.
\bibitem[{Hoerl and Kennard(1970)}]{hoerl_ridge_1970}
\bibinfo{author}{A.~E. Hoerl}, \bibinfo{author}{R.~W. Kennard},
\newblock \bibinfo{title}{Ridge {Regression}: {Biased} {Estimation} for
  {Nonorthogonal} {Problems}},
\newblock \bibinfo{journal}{Technometrics} \bibinfo{volume}{12}
  (\bibinfo{year}{1970}) \bibinfo{pages}{55--67}. \URLprefix
  \url{https://www.jstor.org/stable/1267351}. \DOIprefix\doi{10.2307/1267351},
  \bibinfo{note}{publisher: [Taylor \& Francis, Ltd., American Statistical
  Association, American Society for Quality]}.
\bibitem[{Breiman(2001)}]{breiman_random_2001}
\bibinfo{author}{L.~Breiman},
\newblock \bibinfo{title}{Random {Forests}},
\newblock \bibinfo{journal}{Machine Learning} \bibinfo{volume}{45}
  (\bibinfo{year}{2001}) \bibinfo{pages}{5--32}. \URLprefix
  \url{https://doi.org/10.1023/A:1010933404324}.
  \DOIprefix\doi{10.1023/A:1010933404324}.
\bibitem[{Friedman(2000)}]{friedman_greedy_2000}
\bibinfo{author}{J.~Friedman},
\newblock \bibinfo{title}{Greedy {Function} {Approximation}: {A} {Gradient}
  {Boosting} {Machine}},
\newblock \bibinfo{journal}{The Annals of Statistics} \bibinfo{volume}{29}
  (\bibinfo{year}{2000}). \DOIprefix\doi{10.1214/aos/1013203451}.
\bibitem[{Pedregosa et~al.(2011)Pedregosa, Varoquaux, Gramfort, Michel,
  Thirion, Grisel, Blondel, Prettenhofer, Weiss, Dubourg, Vanderplas, Passos,
  Cournapeau, Brucher, Perrot, and Duchesnay}]{scikit-learn}
\bibinfo{author}{F.~Pedregosa}, \bibinfo{author}{G.~Varoquaux},
  \bibinfo{author}{A.~Gramfort}, \bibinfo{author}{V.~Michel},
  \bibinfo{author}{B.~Thirion}, \bibinfo{author}{O.~Grisel},
  \bibinfo{author}{M.~Blondel}, \bibinfo{author}{P.~Prettenhofer},
  \bibinfo{author}{R.~Weiss}, \bibinfo{author}{V.~Dubourg},
  \bibinfo{author}{J.~Vanderplas}, \bibinfo{author}{A.~Passos},
  \bibinfo{author}{D.~Cournapeau}, \bibinfo{author}{M.~Brucher},
  \bibinfo{author}{M.~Perrot}, \bibinfo{author}{E.~Duchesnay},
\newblock \bibinfo{title}{Scikit-learn: Machine learning in {P}ython},
\newblock \bibinfo{journal}{Journal of Machine Learning Research}
  \bibinfo{volume}{12} (\bibinfo{year}{2011}) \bibinfo{pages}{2825--2830}.
\bibitem[{Liu et~al.(2019)Liu, Ott, Goyal, Du, Joshi, Chen, Levy, Lewis,
  Zettlemoyer, and Stoyanov}]{liu_roberta_2019}
\bibinfo{author}{Y.~Liu}, \bibinfo{author}{M.~Ott}, \bibinfo{author}{N.~Goyal},
  \bibinfo{author}{J.~Du}, \bibinfo{author}{M.~Joshi},
  \bibinfo{author}{D.~Chen}, \bibinfo{author}{O.~Levy},
  \bibinfo{author}{M.~Lewis}, \bibinfo{author}{L.~Zettlemoyer},
  \bibinfo{author}{V.~Stoyanov}, \bibinfo{title}{{RoBERTa}: {A} {Robustly}
  {Optimized} {BERT} {Pretraining} {Approach}}, \bibinfo{year}{2019}.
  \URLprefix \url{http://arxiv.org/abs/1907.11692}.
  \DOIprefix\doi{10.48550/arXiv.1907.11692}, \bibinfo{note}{arXiv:1907.11692
  [cs]}.
\bibitem[{Cañete et~al.(2020)Cañete, Chaperon, Fuentes, Ho, Kang, and
  Pérez}]{CaneteCFP2020}
\bibinfo{author}{J.~Cañete}, \bibinfo{author}{G.~Chaperon},
  \bibinfo{author}{R.~Fuentes}, \bibinfo{author}{J.-H. Ho},
  \bibinfo{author}{H.~Kang}, \bibinfo{author}{J.~Pérez},
\newblock \bibinfo{title}{Spanish pre-trained bert model and evaluation data},
\newblock in: \bibinfo{booktitle}{PML4DC at ICLR 2020}, \bibinfo{year}{2020},
  pp. \bibinfo{pages}{1--10}.
\bibitem[{Wolf et~al.(2020)Wolf, Debut, Sanh, Chaumond, Delangue, Moi, Cistac,
  Rault, Louf, Funtowicz, Davison, Shleifer, von Platen, Ma, Jernite, Plu, Xu,
  Le~Scao, Gugger, Drame, Lhoest, and Rush}]{wolf_transformers_2020}
\bibinfo{author}{T.~Wolf}, \bibinfo{author}{L.~Debut},
  \bibinfo{author}{V.~Sanh}, \bibinfo{author}{J.~Chaumond},
  \bibinfo{author}{C.~Delangue}, \bibinfo{author}{A.~Moi},
  \bibinfo{author}{P.~Cistac}, \bibinfo{author}{T.~Rault},
  \bibinfo{author}{R.~Louf}, \bibinfo{author}{M.~Funtowicz},
  \bibinfo{author}{J.~Davison}, \bibinfo{author}{S.~Shleifer},
  \bibinfo{author}{P.~von Platen}, \bibinfo{author}{C.~Ma},
  \bibinfo{author}{Y.~Jernite}, \bibinfo{author}{J.~Plu},
  \bibinfo{author}{C.~Xu}, \bibinfo{author}{T.~Le~Scao},
  \bibinfo{author}{S.~Gugger}, \bibinfo{author}{M.~Drame},
  \bibinfo{author}{Q.~Lhoest}, \bibinfo{author}{A.~Rush},
\newblock \bibinfo{title}{Transformers: {State}-of-the-{Art} {Natural}
  {Language} {Processing}},
\newblock in: \bibinfo{booktitle}{Proceedings of the 2020 {Conference} on
  {Empirical} {Methods} in {Natural} {Language} {Processing}: {System}
  {Demonstrations}}, \bibinfo{publisher}{Association for Computational
  Linguistics}, \bibinfo{address}{Online}, \bibinfo{year}{2020}, pp.
  \bibinfo{pages}{38--45}. \URLprefix
  \url{https://aclanthology.org/2020.emnlp-demos.6}.
  \DOIprefix\doi{10.18653/v1/2020.emnlp-demos.6}.
\bibitem[{Paszke et~al.(2019)Paszke, Gross, Massa, Lerer, Bradbury, Chanan,
  Killeen, Lin, Gimelshein, Antiga, Desmaison, Kopf, Yang, DeVito, Raison,
  Tejani, Chilamkurthy, Steiner, Fang, Bai, and Chintala}]{NEURIPS2019_9015}
\bibinfo{author}{A.~Paszke}, \bibinfo{author}{S.~Gross},
  \bibinfo{author}{F.~Massa}, \bibinfo{author}{A.~Lerer},
  \bibinfo{author}{J.~Bradbury}, \bibinfo{author}{G.~Chanan},
  \bibinfo{author}{T.~Killeen}, \bibinfo{author}{Z.~Lin},
  \bibinfo{author}{N.~Gimelshein}, \bibinfo{author}{L.~Antiga},
  \bibinfo{author}{A.~Desmaison}, \bibinfo{author}{A.~Kopf},
  \bibinfo{author}{E.~Yang}, \bibinfo{author}{Z.~DeVito},
  \bibinfo{author}{M.~Raison}, \bibinfo{author}{A.~Tejani},
  \bibinfo{author}{S.~Chilamkurthy}, \bibinfo{author}{B.~Steiner},
  \bibinfo{author}{L.~Fang}, \bibinfo{author}{J.~Bai},
  \bibinfo{author}{S.~Chintala},
\newblock \bibinfo{title}{Pytorch: An imperative style, high-performance deep
  learning library},
\newblock in: \bibinfo{booktitle}{Advances in Neural Information Processing
  Systems 32}, \bibinfo{publisher}{Curran Associates, Inc.},
  \bibinfo{year}{2019}, pp. \bibinfo{pages}{8024--8035}. \URLprefix
  \url{http://papers.neurips.cc/paper/9015-pytorch-an-imperative-style-high-performance-deep-learning-library.pdf}.
\bibitem[{Liu et~al.(2023)Liu, Pfeiffer, Vulić, and
  Gurevych}]{liu_improving_2023}
\bibinfo{author}{C.~C. Liu}, \bibinfo{author}{J.~Pfeiffer},
  \bibinfo{author}{I.~Vulić}, \bibinfo{author}{I.~Gurevych},
  \bibinfo{title}{Improving {Generalization} of {Adapter}-{Based}
  {Cross}-lingual {Transfer} with {Scheduled} {Unfreezing}},
  \bibinfo{year}{2023}. \URLprefix \url{http://arxiv.org/abs/2301.05487},
  \bibinfo{note}{arXiv:2301.05487 [cs]}.
\bibitem[{Schmidt et~al.(2021)Schmidt, Goyal, Joshi, Feld, Conell, Laskaris,
  Blank, Wilson, Friedler, and Luccioni}]{codecarbon}
\bibinfo{author}{V.~Schmidt}, \bibinfo{author}{K.~Goyal},
  \bibinfo{author}{A.~Joshi}, \bibinfo{author}{B.~Feld},
  \bibinfo{author}{L.~Conell}, \bibinfo{author}{N.~Laskaris},
  \bibinfo{author}{D.~Blank}, \bibinfo{author}{J.~Wilson},
  \bibinfo{author}{S.~Friedler}, \bibinfo{author}{S.~Luccioni},
\newblock \bibinfo{title}{Codecarbon: estimate and track carbon emissions from
  machine learning computing},
\newblock \bibinfo{journal}{Cited on}  (\bibinfo{year}{2021})
  \bibinfo{pages}{20}.

\end{thebibliography}
\appendix



\section{Dataset Examples} \label{appendix:dataset-examples}

The data was given in JSON format after requesting the server. The following examples are meant to show the structure of how the data was given and later parsed.

\begin{center}
  \begin{minipage}{\linewidth}
\begin{lstlisting}[language=json,firstnumber=1]
[
  {
    "id_message": "1",
    "message": "Me parece que es una buena idea, pero no estoy seguro",
    "date":  "2020-07-27 01:27:31"
  },
  {
      "id_message": 2,
      "message": "Buen dia a todos",
      "date": "2020-07-27 02:03:28"
  },
]
\end{lstlisting}
\captionof{lstlisting}{\textbf{Example of the raw data describing the messages of one user.} The original training set (later split into training+validation) constituted 175 JSON files like this.}
\label{lst:json-example}
\end{minipage}
\end{center}

To complement the files described in the example above, the labels of each subject were given in CSV format, where each row corresponded to one subject. Four of these files were given, one for each task. Table \ref{tab:labels-example} below shows some examples of this. 


\begin{table}[!h]
    \scalebox{0.85}{
    \begin{tabular}{llrlrrrr}
      \toprule
      {} & a\_label &  b\_label & c\_label &  d\_suffer\_in\_favour &  d\_suffer\_against &  d\_suffer\_other &  d\_control \\
      subject\_id &         &          &                   &                   &                 &               &          \\
      \midrule
      subject101 &       1 &      0.9 &  suffer+in favour &               0.7 &             0.1 &           0.1 &      0.1 \\
      subject104 &       1 &      0.7 &  suffer+in favour &               0.4 &             0.0 &           0.3 &      0.3 \\
      subject106 &       1 &      1.0 &  suffer+in favour &               0.5 &             0.5 &           0.0 &      0.0 \\
      subject108 &       1 &      0.5 &  suffer+in favour &               0.4 &             0.1 &           0.0 &      0.5 \\
      subject109 &       0 &      0.1 &           control &               0.0 &             0.0 &           0.1 &      0.9 \\
      \bottomrule
      \end{tabular}
      }
  \caption{\textbf{Example of user labels for each task.} The letter in the prefix of the column name indicates the label of the task (See section \ref{sec:dataset-analysis}). For example, column \textit{a\_label} indicates the label of the users for task 2a.}
  \label{tab:labels-example}
\end{table}

\section{Evaluation of Embedding Models for Regression} \label{appendix:embeddings}

Table \ref{tab:embeddings_rmse} shows the scores after evaluating with different encodings for task 2d. The sentence embeddings were obtained after concatenating the messages of each user into a single string. RMSE scores were calculated as the mean of the results of 10 regressors trained with the respective encoding as features. The best-performing embeddings were the ones obtained with the RoBERTa model fine-tuned for suicide detection.

\begin{table}[h]
    \scalebox{0.9}{
  \begin{tabular}{lcccc|r}
    \toprule
    {} &  suffer+in favour &  suffer+against &  suffer+other &  control &   mean \\
    encoder                               &                   &                 &               &          &        \\
    \midrule
    roberta-base-bne-suicide-es &             0.228 &           0.198 &         0.113 &    0.249 &  0.204 \\
    roberta-base-bne                      &             0.220 &           0.208 &         0.119 &    0.277 &  0.214 \\
    bert-base-spanish-wwm-cased (BETO)           &             0.233 &           0.214 &         0.132 &    0.273 &  0.221 \\
    \bottomrule
    \end{tabular}
    }
    \caption{
      Mean RMSEs scores after training 10 regressors with embeddings from different transformers. 
    }
    \label{tab:embeddings_rmse}
\end{table}

The difference in performance between \texttt{roberta-base-bne-suicide-es} encodings and the other embeddings can be justified by the fact that we are taking advantage of the information gained from the prior fine-tuning for suicide detection of this model, which likely shares semantic similarities with our data.

The implementation of the sentence embeddings was done using the HuggingFace library. Before obtaining the encodings, the models were loaded from the \href{https://huggingface.co/models}{HuggingFace Hub}.  The reference and links to the models in Hugging Face is included in \ref{tab:bert_models}.

\section{Evaluation of Regression models trained with the Sentence Embeddings Approach} \label{appendix:embedding-regressor-eval}


Tables \ref{tab:eval_results_2b_embeddings} and \ref{tab:results_2d_chainpca} report the regression metrics obtained on the validation set for various estimators trained with the \texttt{roberta-base-bne-suicide-es} embeddings. The results were obtained after evaluating over 10 estimators for the task, selecting the 4-6 best-performing regressors, and doing grid-search hyperparameter-tuning on them. 

By these results, the Ridge Regression model (ridge), which implements Least Squares Linear Regression with L2 regularization, seems to be the best estimator for both tasks. The best-performing models of each task were then trained with the entire data (train+validation sets) and the prediction of the test set was obtained with them. These evaluations with these predictions are the ones reported in section \ref{sec:results}.

\begin{table}[h]
  \centering
  \vspace*{-0.2cm}
  \caption{Binary regression results in the evaluation set of various estimators trained with embeddings for task 2b.
  }
  \label{tab:eval_results_2b_embeddings}
  \begin{tabular}{lr|r}
  \toprule
  {} &   rmse &     r2 \\
  estimator &        &        \\
  \midrule
  ridge     &  0.254 &  0.489 \\
  ada       &  0.257 &  0.480 \\
  lgbm      &  0.259 &  0.471 \\
  svr       &  0.260 &  0.467 \\
  rf        &  0.264 &  0.448 \\
  mlp       &  0.319 &  0.198 \\
  \bottomrule
  \end{tabular}
\end{table}

\begin{table}[h]
  \centering
  \caption{Multi-output regression results in the evaluation set of various estimators trained with embeddings for task 2d. Labels are shortened as:
   sf = suffer+in favour, sa = suffer+against, so = suffer+other, c = control}
  \label{tab:results_2d_chainpca}
  \scalebox{0.85}{
  \begin{tabular}{lr|rrrrrrrrr}
  \toprule
  {} &  RMSE mean &  RMSE sf &  RMSE sa &  RMSE so &  RMSE c &    R2 mean &  R2 sf &  R2 sa &  R2 so &  R2 c \\
  estimator &            &                        &                      &                    &               &            &                      &                    &                  &             \\
  \midrule
  ridge     &      0.183 &                  0.172 &                0.193 &              0.108 &         0.260 &      0.271 &                0.410 &              0.287 &           -0.076 &       0.465 \\
  lr        &      0.193 &                  0.202 &                0.197 &              0.108 &         0.267 &      0.201 &                0.183 &              0.256 &           -0.073 &       0.439 \\
  rf        &      0.206 &                  0.195 &                0.220 &              0.106 &         0.305 &      0.133 &                0.240 &              0.071 &           -0.045 &       0.267 \\
  lgbm      &      0.288 &                  0.285 &                0.262 &              0.181 &         0.424 &     -0.854 &               -0.631 &             -0.321 &           -2.043 &      -0.421 \\
  \bottomrule
  \end{tabular}
  }
\end{table}

The estimators mentioned in the table above are implementations of common regressors from Python's Scikit-Learn library \cite{scikit-learn}. These include: Ordinary ("lr") and Ridge ("ridge") Least Squares Regression, Ada-Boost regression ("ada"), Light Gradient Boosting Machine ("lgbm"), Support Vector Regression ("svr"), Random Forests ("rf"), and a Multi-Layer Perceptron ("mlp"). References of the implementations of these models can be found in the \href{https://scikit-learn.org/stable/supervised_learning.html}{Scikit-Learn documentation}.

\section{Multi-Output Regression with Independent Regressors and Regressor Chains} \label{appendix:multi-output-regression}

For task 2d, we were required to obtain four values corresponding to a probability distribution over the four classes (suffer+in favour, suffer+against, suffer+other, control). In section \ref{sec:solving-for}, we explained how this multi-output regression problem can be solved for the sentence embeddings approach by training four regressors and then combining their predictions using either the Independent Regressors or Regressor Chain methods. 

Here we explain how these methods work and were implemented. First of all, the two methods can be summarized as follows, depending on how the regressors are trained to obtain the probability distributions. Figure \ref{fig:multi-regression-methods} shows a graphical representation of the two methods.

\begin{enumerate}
  \item \textbf{Independent Regressors}: Each regressor is trained independently with the labels of its corresponding class (e.g., the first regressor was trained with the labels of the suffer+in favour class, the second with the labels of the suffer+against class, and so on). The downside of this method is that it doesn't take into account the information of the other classes when training each regressor, which is important as we know the labels are not independent of each other (they must all sum to 1).
  \item \textbf{Regressor Chain Method}: The regressors are trained in a chain, where the first regressor is trained to predict the first class, and its predictions are included in the features for the second regressor, and so on. This method is useful when the labels of each class are not independent of each other (like in our case), as it allows the regressors to learn from the predictions of the previous ones. Since the order of the classes matters in this method, we decided to put them in the order of most to least amount of users annotated with that class: control, suffer+in favour, suffer+against, suffer+other (see section \ref{sec:dataset-analysis}).
\end{enumerate}

To the second method, we can additionally add the option of applying Principal Component Analysis (PCA) to reduce the dimensionality of the input embeddings before training the models in the chain. Because the embeddings might have a large dimensionality, this is done to make these models more likely to use the information of the previous predictions.
Both methods were implemented with the Scikit-Learn library \cite{scikit-learn} using the \texttt{MultiOutputRegressor} and \texttt{RegressorChain} classes. The number of components to keep for PCA was chosen using based on the percent of variance explained. The number of components was fine-tuned and the best results was obtained with 40 components (85\% of variance).


\end{document}